\documentclass[letterpaper, 10 pt, journal, twoside]{IEEEtran}
\ifCLASSINFOpdf
\else
\fi
\usepackage{url}

\usepackage{graphics} 
\usepackage{epsfig} 
\usepackage{mathptmx} 
\usepackage{times} 
\usepackage{amsmath} 
\usepackage{amssymb}  

\usepackage[dvipsnames]{xcolor}
\usepackage{algorithm}
\usepackage{algorithmicx}
\usepackage{algpseudocode}
\usepackage{amsmath}
\usepackage{bm}

\usepackage{booktabs}
\usepackage{subcaption}
\usepackage{hyperref}
\definecolor{brightturquoise}{rgb}{0.03, 0.91, 0.87}
\usepackage{multirow}
\usepackage{array}

\begin{document}
%
\title{Cross-view Semantic Segmentation for Sensing Surroundings}
%
%
%

\author{Bowen Pan$^{1,*}$, Jiankai Sun$^{2,*}$, Ho Yin Tiga Leung$^{2}$, Alex Andonian$^{1}$, and Bolei Zhou$^{2}$\thanks{* indicates equal contribution.}\thanks{Manuscript received: February 24, 2020; Revised May 13, 2020; Accepted June 4, 2020.}\thanks{This paper was recommended for publication by Editor Tamim Asfour upon evaluation of the Associate Editor and Reviewers' comments.
This work was supported by CUHK FoE Direct Grant and Facebook PyRobot Research Award.}\thanks{$^{1}$ B. Pan and A. Andonian are with the Computer Science and Artificial Intelligence Laboratory, Massachusetts Institute of Technology, USA.}\thanks{$^{2}$ J. Sun, H. Y. T. Leung and B. Zhou are with the Department of Information Engineering, The Chinese University of Hong Kong, Hong Kong, China. Corresponding email: {\tt\footnotesize bzhou@ie.cuhk.edu.hk}}\thanks{Digital Object Identifier (DOI): see top of this page.}}
%
%

\markboth{IEEE Robotics and Automation Letters. Preprint Version. Accepted June, 2020}
{Pan \MakeLowercase{\textit{et al.}}: Cross-view Semantic Segmentation for Sensing Surroundings}

%



\maketitle
\begin{abstract}
Sensing surroundings plays a crucial role in human spatial perception, as it extracts the spatial configuration of objects as well as the free space from the observations. To facilitate the robot perception with such a surrounding sensing capability, we introduce a novel visual task called Cross-view Semantic Segmentation as well as a framework named View Parsing Network (VPN) to address it. In the cross-view semantic segmentation task, the agent is trained to parse the first-view observations into a top-down-view semantic map indicating the spatial location of all the objects at pixel-level. The main issue of this task is that we lack the real-world annotations of top-down-view data. To mitigate this, we train the VPN in 3D graphics environment and utilize the domain adaptation technique to transfer it to handle real-world data. We evaluate our VPN on both synthetic and real-world agents. The experimental results show that our model can effectively make use of the information from different views and multi-modalities to understanding spatial information. Our further experiment on a LoCoBot robot shows that our model enables the surrounding sensing capability from 2D image input. Code and demo videos can be found at \url{https://view-parsing-network.github.io}.
\end{abstract}

\begin{IEEEkeywords}
Semantic Scene Understanding, Deep Learning for Visual Perception, Visual Learning, Visual-Based Navigation, Computer Vision for Other Robotic Applications
\end{IEEEkeywords}

%
\IEEEpeerreviewmaketitle

\section{Introduction}
%
%
%
%
\IEEEPARstart{R}{ecent} progress in semantic understanding enables machine perception to segment a scene precisely into meaningful regions and objects \cite{ kostavelis2015semantic, sunderhauf2016place}.
These semantic segmentation techniques have benefited many automation applications, like autonomous driving \cite{cordts2016cityscapes}.
Though the semantic segmentation network can recognize semantic content in a static image, it is still far from enough to facilitate robots to sense in an unknown environment and navigate freely there. One important reason is that the parsed first-view semantic mask is still at pure image-level without providing any spatial information about the surroundings. To perceive spatial configuration from pure image input, an intuitive approach is to explicitly train networks to infer the top-down-view semantic map which directly contains the spatial configuration information of the surrounding environment. Based on the top-down-view semantic map we can then infer the position coordinates and functional properties of surrounding regions and objects.

\begin{figure}
\begin{center}
\includegraphics[width=0.9\linewidth]{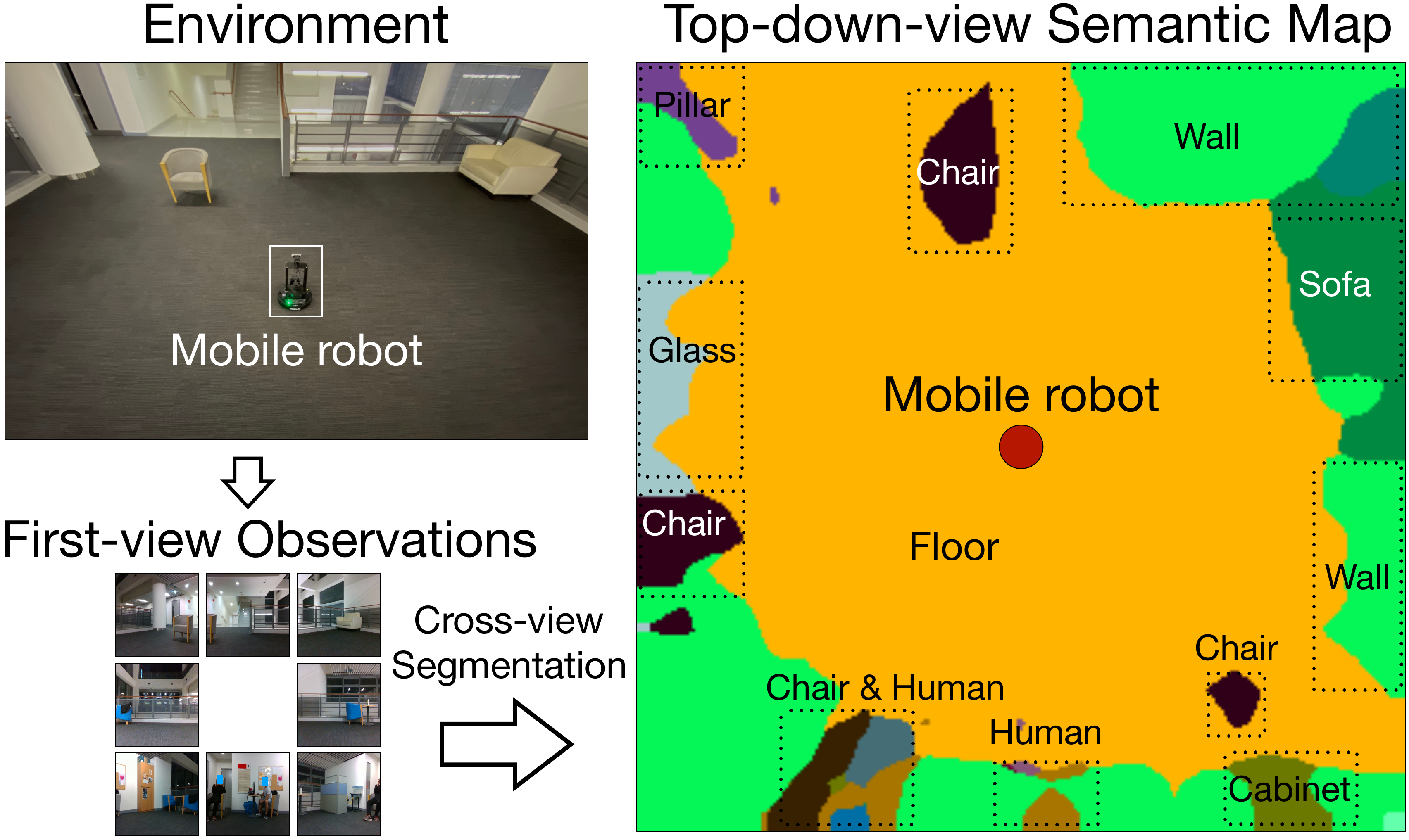}
\end{center}
   \caption{Top-down-view semantics is predicted from the first-view real-world observations in the cross-view semantic segmentation. Input observations from multiple angles are fused. Notice that the result in this figure is generated without training on real-world data. 
   }
\label{fig:cover}
\end{figure}

To enable machines to capture the spatial structure of the surroundings from 2D images, we explore a new image-based scene understanding task, \textit{Cross-View Semantic Segmentation}. Different from the standard semantic segmentation predicting the labels of each pixel in the input image, the cross-view semantic segmentation aims at predicting the top-down-view semantic map from a set of first-view observations (see Fig.~\ref{fig:cover}). The resulting top-down-view semantic map, as a 2.5D spatial representation of the surrounding, indicates the spatial layout of the discrete objects such as chair and human, as well as the stuff classes floor and wall. Note that although there is a huge literature of 3D methods to reconstruct environments~\cite{dai2017scannet}, our method has its unique advantages. For example, robot perception systems based on 3D sensors involve expensive cost not only in sensor setup but also in computational power. Instead, the top-down-view map from the cross-view semantic segmentation can facilitate the robot to understand its surroundings in a lightweight and efficient way. In many situations such as free space exploration for mobile robots where the height information is not that essential, the 2D top-down-view semantic map would be sufficient to provide spatial information with much less computation cost.

One challenge in cross-view semantic segmentation is the difficulty of collecting the top-down-view semantic annotations. Recently, simulation environments such as House3D \cite{wu2018building} and CARLA \cite{Dosovitskiy17} have been proposed for training navigation agents. In these environments, cameras can be placed at any location in the simulated scene while the observations in multiple modalities can be extracted. Thus, we leverage the simulation environments to acquire cross-view annotated data.
To reduce the domain gap between the synthetic scenes and the real-world scenes, we transfer the models trained in the simulation environment to the real-world scenes through domain adaptation.

In this work, we propose a novel framework with View Parsing Network (VPN) for cross-view semantic segmentation using simulation environments and then transfer them to real-world environments. In VPN, a view transformer module is designed to aggregate the information from multiple first-view observations with different angles and different modalities. It outputs the top-down-view semantic map with a spatial layout of objects. 
We evaluate the proposed models on the indoor scene of the House3D environment \cite{wu2018building} and the outdoor driving scene of the CARLA environment \cite{Dosovitskiy17}.
Furthermore, to show the cross-view semantic task helps visual navigation, we have demonstrations of real robot. 



Our main contributions are as follows: (1) We introduce a novel task named \emph{cross-view semantic segmentation} to facilitate robots to flexibly sense the surrounding environment. (2) We propose a framework with \emph{View Parsing Network} which effectively learns and aggregates features across first-view observations with multiple angles and modalities. (3) We further apply the domain adaptation technique to transferring our model so that it can work in real-world data while without any extra annotations.

\begin{figure*}
\centering
\includegraphics[width=\linewidth]{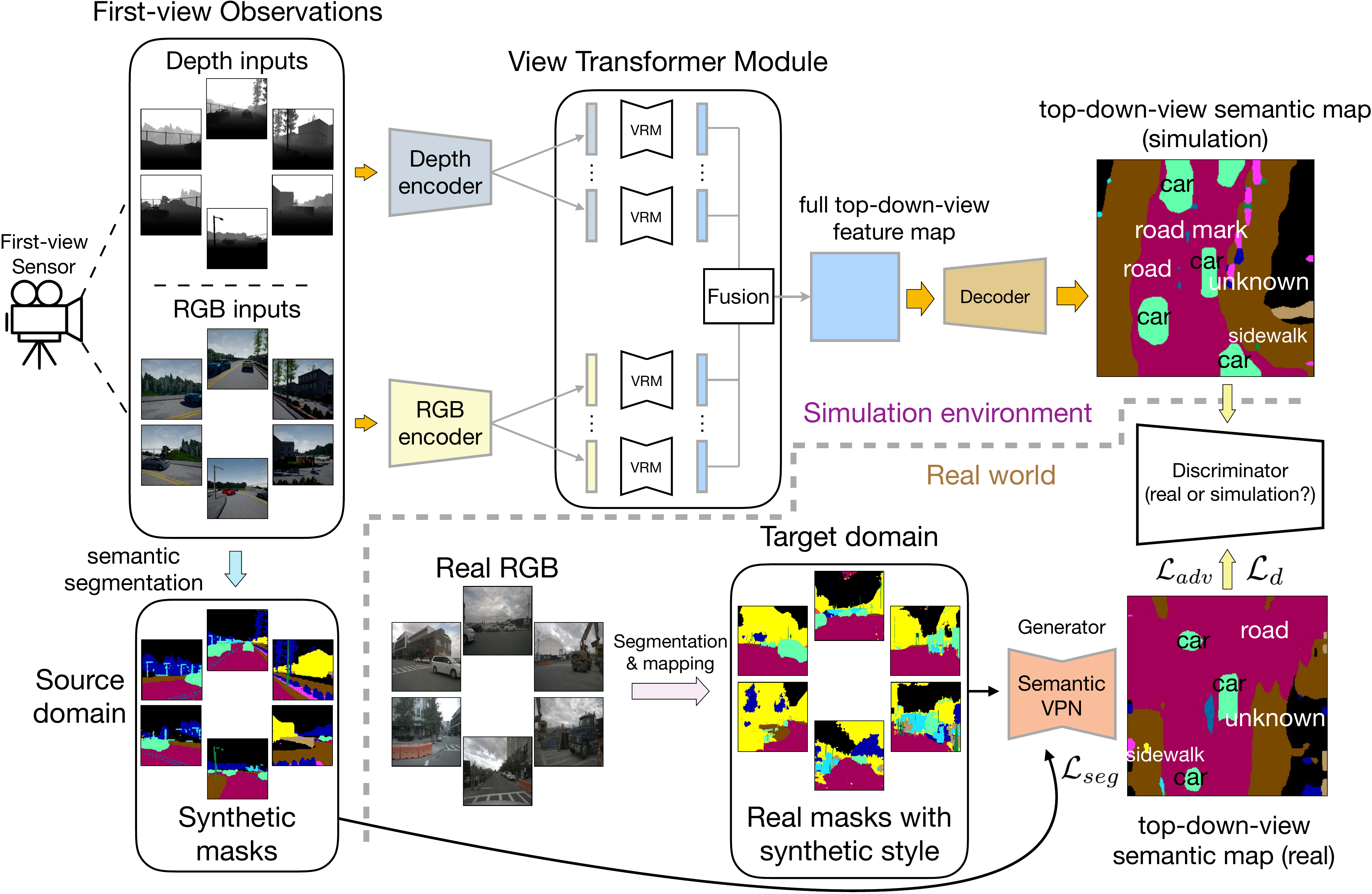}
\caption{Framework of the View Parsing Network for cross-view semantic segmentation. The simulation part shows the architecture and training scheme of our VPN, while the real-world part demonstrates the domain adaptation process for transferring VPN to the real world.}
\label{fig:framework}
\end{figure*}

%

\section{RELATED WORK}
\subsection{Semantic Segmentation and Semantic Mapping}
Deep learning networks for semantic segmentation \cite{zhao2017pspnet} 
are designed to segment the image pixel-wise within one-view. Image datasets with pixel-wise annotations such as CityScapes \cite{cordts2016cityscapes} are used for the training of semantic segmentation networks.
There is also a huge literature about semantic mapping in robotics domain~\cite{kostavelis2015semantic, sunderhauf2016place, katsumata2019semantic, zheng2018learning}, which provides the semantic abstraction of the environment and a way to communicate with robots.

\subsection{Layout estimation and view synthesis}
Estimating layout has been an active topic of research (i.e. room layout estimation \cite{zou2018layoutnet}, free space estimation \cite{hedau2012recovering}, \
and road layout estimation \cite{schulter2018learning, wang2019parametric}). Most of the previous methods use annotations of the layout or geometric constraints for the estimation, while our proposed framework estimates the top-down-view map directly from the image, without the intermediate step of estimating the 3D structure of the scene. On the other hand, view synthesis has been explored in many works \cite{zhai2017predicting,regmi2018cross}. They focus on generating realistic cross-view images while cross-view segmentation aims at parsing semantics across different views. 

\subsection{Learning in Simulation Environments}
Given that current graphics simulation engines can render realistic scenes, recognition algorithms can be trained on data pulled from simulation engines (\emph{i.e.,} for visual navigation models \cite{shen2019situational}). Several techniques have been proposed to address the domain adaptation issue when models trained with simulated images are transferred to real scenes \cite{tsai2018learning}. 
Rather than working on the task of visual navigation directly, our work aims at parsing the top-down-view semantic map from the first-view observations. The resulting top-down-view map will further facilitate visual navigation.

\section{CROSS-VIEW SEMANTIC SEGMENTATION}

\subsection{Problem Formulation} \label{sec:formulation}
The objective of cross-view semantic segmentation is as follows: given the first-view observations as input, the algorithm must generate the top-down-view semantic map. 
The top-down-view semantic map is a map captured by a camera at a certain height from the top-down view with the annotations of the semantic label of each pixel.
The input first-view observations are a set of images with different modalities. They are captured at $N$ different angles by the robot's camera (with $360 / N$ degrees apart). 


\subsection{Framework of the View Parsing Network}\label{framework}

Fig.~\ref{fig:framework} illustrates two stages of our framework. In the first stage, we propose View Parsing Network (VPN) to learn and aggregate features from multiple first-view observations in the simulation environment. In VPN, first-view observations are first fed into the encoder to extract first-view feature maps. For each modality, VPN has a corresponding encoder to process it. All of these first-view feature maps from different angles and different modalities are transformed and then aggregated into one top-down-view feature map in the View Transformer Module. Then the aggregated feature map is decoded into a top-down-view semantic map. Details of how to transform and aggregate these first-view feature maps can be found in Sec.~\ref{exp:arch}. In the second stage of our framework, we transfer the knowledge which VPN learns from the simulation environment to the real-world data. We slightly modified the domain adaptation algorithm proposed by \cite{tsai2018learning} to fit our cross-view semantic segmentation task and our VPN architecture. More details of this part will be revealed in Section~\ref{sec:syn2real}.

\emph{Pipeline.}
As shown in Fig.~\ref{fig:framework}, from one spatial position in a 3D environment, we first sample $N\times M$ first-view observations from $N$ angles and $M$ modalities (here $N=6, M=2$ in Fig.~\ref{fig:framework}) in even angles so that all-around information is captured. 
The first-view observations are encoded by $M$ encoders for $M$ corresponding modalities respectively. 
These CNN-based encoders extract $N\times M$ spatial feature maps for their first-view input. Then all of these feature maps are fed into the View Transformer Module (VTM). VTM transforms these view feature maps from first-view space into the top-down-view feature space and fuses them to get one final feature map which already contains sufficient spatial information. Finally, we decode it to predict the top-down-view semantic map using a convolutional decoder.

\emph{View Transformer Module.}
Although the encoder-decoder structure gets huge success in the classical semantic segmentation area \cite{
zhao2017pspnet}, our experiment (cf. Table~\ref{tab-vpn-abl}) shows that it performs poorly in the cross-view semantic segmentation task. We conjecture that it is because in standard semantic segmentation architecture the receptive field of the output spatial feature map is roughly aligned with the input spatial feature map. However, in cross-view semantic segmentation, each pixel on the top-down-view map should consider all input first-view feature maps, not just a local receptive field region. 

After thinking about the flaws of the current semantic segmentation structure, we design the View Transformer Module (VTM) to learn the dependencies across all the spatial locations between the first-view feature map and the top-down-view feature map.
VTM will not change the shape of input feature map, so it can be plugged into any existing encoder-decoder type of network architecture for classical semantic segmentation. It consists of two parts: View Relation Module (VRM) and View Fusion Module (VFM). The diagram at the central of Figure~\ref{fig:framework} illustrates the whole process: The first-view feature map is first flattened while the channel dimension remains unchanged. Then we use a view relation module $R$ to learn the relations between the any two pixel positions in flattened first-view feature map and flattened top-down-view feature map. That is:
\vspace{-0.1cm}
\begin{equation}
    f_t[i] = R_i(f[1], ..., f[j], ..., f[HW]),
    \vspace{-0.1cm}
    \label{eq:vrm}
\end{equation}
where $i, j\in [0, HW)$ are the indices of top-down-view feature map $t\in R^{HW\times C}$ and first-view feature map $f\in R^{HW\times C}$ respectively along the flattened dimension, and $R_i$ models the relations between the $i^{th}$ pixel on top-down-view feature map and every pixel on first-view feature map. Here we simply use multilayer perceptron (MLP) in our view relation module $R$. After that, the top-down-view feature map is reshaped back to $H \times W \times C$. Notice that each first-view input has its own VRM to get the top-down-view feature map $t^i\in R^{H\times W \times C}$ based on its own observations. To aggregate the information from all observation inputs, we fuse these top-down-view feature map $t^i$ by using VFM. More details of VFM and VRM will be introduced in Sec.~\ref{exp:arch}.


\begin{table*}[!htb]
\centering
\small
\caption{Results on House3D cross-view dataset with different modalities and view numbers.}
\begin{tabular}{ccccccccccc}
    \toprule
     &  \multicolumn{2}{c}{3D Geometric Baseline}& \multicolumn{2}{c}{X-Fork(RGB) in \cite{regmi2018cross}}&  \multicolumn{2}{c}{RGB VPN} & \multicolumn{2}{c}{Semantic VPN} & \multicolumn{2}{c}{Depth VPN} \\
    \cmidrule(lr){2-3} \cmidrule(lr){4-5} \cmidrule(lr){6-7} \cmidrule(lr){8-9} \cmidrule(lr){10-11}
    Networks & PA & mIoU &  PA & mIoU & PA & mIoU &  PA &  mIoU &  PA &  mIoU \\
    \cmidrule(lr){1-1}     \cmidrule(lr){2-3} \cmidrule(lr){4-5} \cmidrule(lr){6-7} \cmidrule(lr){8-9} \cmidrule(lr){10-11}
      1-view model & $31.3\%$  & $2.4\%$ & $38.0\%$ & $1.5\%$ & $55.8\%$ &  $6.5\%$ &  $59.6\%$  &  $13.2\%$ & $56.9\%$ & $7.6\%$ \\
    \cmidrule(lr){1-1}     \cmidrule(lr){2-3} \cmidrule(lr){4-5} \cmidrule(lr){6-7} \cmidrule(lr){8-9} \cmidrule(lr){10-11}
    2-view model & $46.8\%$  & $7.2\%$ & $40.0\%$ & $1.9\%$ & $70.1\%$ & $14.8\%$ & $75.7\%$ & $25.9\%$ & $70.2\%$ & $15.6\%$ \\
        \cmidrule(lr){1-1}     \cmidrule(lr){2-3} \cmidrule(lr){4-5} \cmidrule(lr){6-7} \cmidrule(lr){8-9} \cmidrule(lr){10-11}
      4-view model & $63.2\%$  & $22.8\%$ & $39.5\%$ & $2.0\%$ &  $80.3\%$ &  $27.2\%$ &   $\bm{85.0\%}$ &    $40.6\%$ &    $77.3\%$ &  $22.0\%$ \\
    \cmidrule(lr){1-1}     \cmidrule(lr){2-3} \cmidrule(lr){4-5} \cmidrule(lr){6-7} \cmidrule(lr){8-9} \cmidrule(lr){10-11}
     8-view model & $67.6\%$  & $27.1\%$ & $43.9\%$ & $1.8\%$ & $\bm{81.2\%}$ & $\bm{28.5\%}$ & $84.7\%$ & $\bm{41.0\%}$ & $\bm{82.1\%}$ & $\bm{29.9\%}$ \\
    \bottomrule
\end{tabular}    
    \label{tab-vpn}
\end{table*}

\begin{table}
\centering
\small
\caption{Results of cross-modality learning for VPN. Here we compare the results with the inputs from 4 views.
}
\begin{tabular}{ccc}
    \toprule
     Method &  Pixel Accuracy & mIoU \\
    \cmidrule(lr){1-1} \cmidrule(lr){2-2} \cmidrule(lr){3-3}
    RGB VPN & $80.3\%$ &  $27.2\%$ \\
    \cmidrule(lr){1-1} \cmidrule(lr){2-2} \cmidrule(lr){3-3}
    Depth VPN &  $77.3\%$ & $22.0\%$ \\
    \cmidrule(lr){1-1} \cmidrule(lr){2-2} \cmidrule(lr){3-3}
    Semantic VPN & $85.0\%$ & $40.6\%$ \\
     \hline
     \hline
     R+D (late fusion) & $81.2\%$  & $27.3\%$  \\
     R-D VPN & $82.8\%$ &  $31.2\%$ \\
    \cmidrule(lr){1-1} \cmidrule(lr){2-2} \cmidrule(lr){3-3}
    D+S (late fusion) & $83.5\%$ & $33.9\%$  \\
    D-S VPN & $\bm{86.3\%}$  & $43.2\%$ \\
    \cmidrule(lr){1-1} \cmidrule(lr){2-2} \cmidrule(lr){3-3}
    S+R (late fusion) & $84.3\%$ & $35.7\%$ \\
    S-R VPN & $85.1\%$ & $42.3\%$ \\
    \cmidrule(lr){1-1} \cmidrule(lr){2-2} \cmidrule(lr){3-3}
    D+S+R (late fusion) & $84.3\%$ & $29.4\%$ \\
    D-S-R VPN & $86.2\%$ & $\bm{43.6\%}$ \\
\bottomrule
\end{tabular}    
    \label{tab:vpn-cross}
\end{table}

\subsection{Sim-to-real Adaptation}\label{sec:syn2real}
To generalize our VPN to real-world data without the real-world ground truth, we implement the sim-to-real domain adaptation scheme shown in  Fig.~\ref{fig:framework} to narrow the gap. This scheme contains the following pixel-level adaptation and output space adaptation.

\emph{Pixel-level adaptation.}
To mitigate the domain shift, we adopt the pixel-level adaptation on the real-world inputs to make them look more like the style of the simulation data. 
Semantic mask is an ideal mid-level representation without texture gap while including sufficient information and it is easy to transfer. 
This process can be formulated as follows:
\begin{equation}
    \{I_S\} = \mathcal{M}_{Real\rightarrow Synthetic}(\mathcal{P}_{RGB\rightarrow Mask}(\{I_R\})),
    \label{eq:adaptpix}
\end{equation}
where $I_R$, $I_S$ are the real RGB image and synthetic-style semantic mask respectively, $\mathcal{P}_{RGB\rightarrow Mask}$ is the existing semantic segmentation model which parses the real-world RGB into semantic mask, and $\mathcal{M}_{Real\rightarrow Synthetic}$ is the semantic category mapping process where we construct the concept mappings between the real world and the simulation environment.

\emph{Output space adaptation.}
Beyond the pixel-level transfer on input data, we also devise an adversarial training scheme in structured output space based on the method proposed in \cite{tsai2018learning}. Here the generator $\mathcal{G}$ is a view parsing network generating the top-down-view prediction $P$,
which is initialized by the weights of a VPN trained on the semantic data in the simulation environment as we illustrated before. During the training phase, we first forward a group of input images from the source domain $\{I_s\}$ to $\mathcal{G}$ and optimize it with a normal segmentation loss $\mathcal{L}_{seg}$. Then we use $\mathcal{G}$ to extract the feature map $F_i$ (after the softmax layer) of the images from the target domain $\{I_t\}$ and use discriminator to distinguish whether $F_t$ is from the source domain. 
The loss function to optimize $\mathcal{G}$ can be written as follows:
\begin{equation}
    \mathcal{L}(\{I_s\}, \{I_t\}) = \mathcal{L}_{seg}(\{I_s\}) + \lambda_{adv}\mathcal{L}_{adv}(\{I_t\}),
    \label{eq:adaptloss}
\end{equation}
where $\mathcal{L}_{seg}$ is the cross-entropy loss for semantic segmentation, $\mathcal{L}_{adv}$ is designed to train the $\mathcal{G}$ and fool the discriminator $\mathcal{D}$. The loss function for the discriminator $\mathcal{L}_d$ is a cross-entropy loss for binary source \& target classification.

\subsection{Network configuration}\label{exp:arch}
\emph{View encoder and decoder.}
To balance efficiency and performance, we use ResNet-18 as the encoder. We remove the last Residual Block and the Average Pool layer so that the resolution of the encoding feature map remains large, which better preserves the details of the view. We employ the pyramid pooling module used in \cite{zhao2017pspnet} as the decoder. 

\emph{View Transformer Module.}
For each view relation module, we simply use the two-layer MLP. We choose this because two-layer MLP doesn't bring too much extra computation so that we can keep our model following the lightweight-and-efficient rationale. Input and output dimensions of the VRM are both $H_IW_I$, where $H_I$ and $W_I$ are respectively the height and width of the intermediate feature map. 
As for the view fusion module, we just add all the features up to keep the shape consistent. 

\emph{Sim-to-real.}
For the generator $\mathcal{G}$, we use the architecture of the 4-view VPN. For the discriminator $\mathcal{D}$, we adopt the same architecture in~\cite{tsai2018learning}. It has 5 convolution layers, each of which is followed by a leaky ReLU with the parameter 0.2 (except the last layer). We use HRNet~\cite{sun2019high} pretrained on CityScapes dataset~\cite{cordts2016cityscapes} to extract the semantic mask from real-world images.

\section{EXPERIMENTS}\label{sec:exp}
We first go through the overview of the cross-view segmentation datasets in Section~\ref{exp:dataset}. Then we show the performance of VPN on synthetic data of the House3D and CARLA environment in Section~\ref{eval}. Finally in Section~\ref{sec:sim2real}, we demonstrate the real-world performance of our VPN which is trained in the simulation environment. 

\begin{table}
\centering
\small
\caption{Ablation study of View Transformer Module.
}
\begin{tabular}{ccccc}
    \toprule
     Modality &  \multicolumn{2}{c}{VPN w/o VTM} & \multicolumn{2}{c}{VPN} \\
    \hline
    1-view & Pix. Acc. & mIoU & Pix. Acc. & mIoU \\
    \cmidrule(lr){1-1} \cmidrule(lr){2-3} \cmidrule(lr){4-5}
     RGB & $53.9\%$ &  $6.3\%$ & $\bm{55.8\%}$ &  $\bm{6.5\%}$ \\
    \cmidrule(lr){1-1} \cmidrule(lr){2-3} \cmidrule(lr){4-5}
    Depth &  $55.7\%$ & $6.5\%$ &  $\bm{56.9\%}$ & $\bm{7.6\%}$ \\
    \cmidrule(lr){1-1} \cmidrule(lr){2-3} \cmidrule(lr){4-5}
    Semantic & $57.4\%$ & $10.0\%$ & $\bm{59.6\%}$ & $\bm{13.2\%}$ \\
     \hline
     \hline
    8-view & Pix. Acc. & mIoU & Pix. Acc. & mIoU \\
    \cmidrule(lr){1-1} \cmidrule(lr){2-3} \cmidrule(lr){4-5}
     RGB & $60.5\%$ &  $8.7\%$ & $\bm{81.2\%}$ &  $\bm{28.5\%}$ \\
    \cmidrule(lr){1-1} \cmidrule(lr){2-3} \cmidrule(lr){4-5}
    Depth &  $43.8\%$ & $2.5\%$ &  $\bm{82.1\%}$ & $\bm{29.9\%}$ \\
    \cmidrule(lr){1-1} \cmidrule(lr){2-3} \cmidrule(lr){4-5}
    Semantic & $47.6\%$ & $6.5\%$ & $\bm{84.7\%}$ & $\bm{41.0\%}$ \\
\bottomrule
\end{tabular}    
    \label{tab-vpn-abl}
\end{table}


\subsection{Benchmarks}\label{exp:dataset}
Here we introduce \emph{two synthetic cross-view datasets}, \emph{House3D cross-view dataset} and \emph{Carla cross-view dataset}, and \emph{one real-world cross-view dataset}, \emph{nuScenes dataset}. 

\emph{House3D cross-view dataset.} 
Each data pair contains 8 first-view input images captured from 8 different orientations with 45 degrees apart.  Additionally, each data pair comes with the top-down-view semantic mask captured in the ceiling-level height. To be complete, we store the input image with multiple modalities including the RGB images, depth maps, and semantic masks. 
The training set contains 143k data pairs from 342 scenes while the validation set contains 20k data pairs from 68 scenes.

\emph{NuScenes dataset.} 
Each data sample contains in NuScenes\cite{nuscenes2019} first-view RGB images from 6 directions (\emph{Front, Front-right, Back-right, Back, Back-left, Front-left}) in different modalities. 
We select 919 data samples without the top-down-view mask for unsupervised training and 515 data samples with the binary top-down-view mask for evaluation.

\emph{CARLA cross-view dataset.}
To build the synthetic source domain dataset, we extract $28,000$ data pairs with top-down-view annotations and different input modalities from $14$ driving episodes in CARLA. Each data pair contains 6 first-view input image sets captured from the same 6 directions.


\subsection{Evaluation}\label{eval}

We present VPN performances on the synthetic data of House3D cross-view and CARLA cross-view datasets. 

\emph{Metrics.}
We report the results of cross-view semantic segmentation using two commonly used metrics in semantic segmentation: \textsc{Pixel Accuracy (PA)} which characterizes the proportion of correctly classified pixels, and \textsc{Mean IoU (mIoU)} which indicates the intersection-and-union between the predicted and ground truth pixels.

\emph{Baselines.}
Two methods are included as the comparison baselines: (1) \emph{3D geometric method}. With the observed depth and RGB  images, we can reconstruct the 3D points cloud with the voxel-level semantic label. 
(2) \emph{Cross-view synthesis.} We also compare with the architecture used in cross-view image synthesis literature~\cite{regmi2018cross}, which adopts a conditional GAN called X-Fork to generate aerial images from street-view images. 

\subsubsection{Results of VPNs} We present the results of our VPN for cross-view semantic segmentation in House3D, including the ones of single-modality and multi-modalities VPN respectively. To better evaluate our VPN, we impose an upper bound that we perform segmentation using top-view RGB images directly as inputs, where we get the performance of \emph{$91.4\%$ pixel acc.} and \emph{$41.2\%$ mIoU.} We also show the comparison with the geometric baseline and the ablation study of View Transformer Module. 

\begin{figure}
\centering
\includegraphics[width=\linewidth]{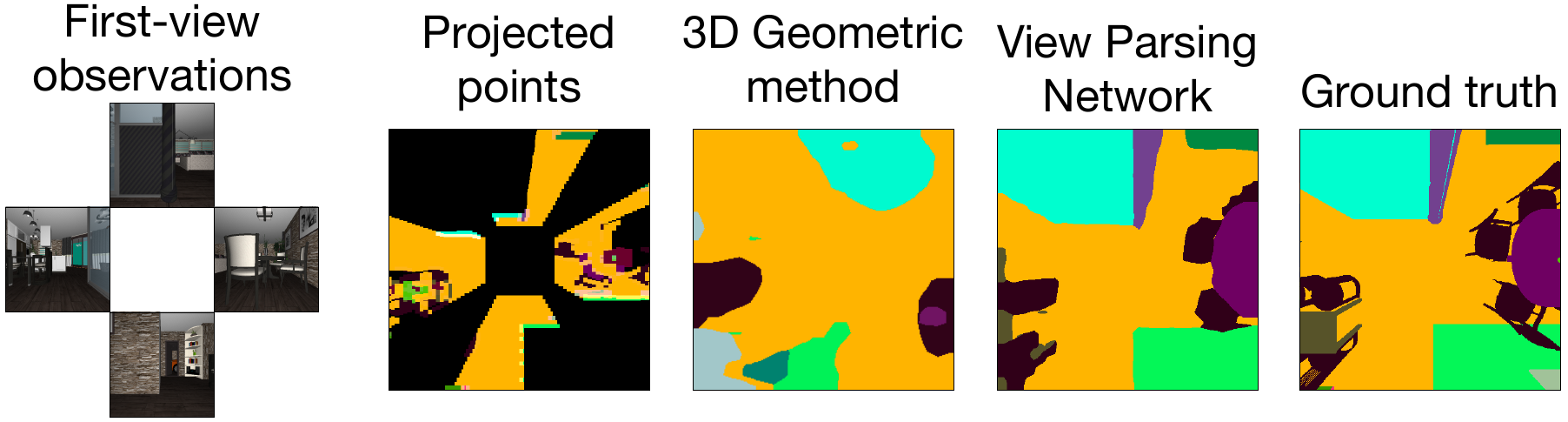}
\caption{Qualitative results of 3D geometric method and our VPN. Considering that the geometric method requires semantic mask and depth map, we use the 4-view Depth-Semantic VPN to predict the top-down-view semantic map to fairly compare these two methods.}
\label{fig:3d_vpn}
\end{figure}

\emph{Single-modality VPN.} We show the House3D results of single-modality VPN with different modalities and different numbers of views in Table~\ref{tab-vpn}. We can see that as VPN receives more views, the segmentation results improve rapidly. 
We also plot some qualitative results by our VPNs in Fig.~\ref{fig:3d_vpn}. On Carla dataset, we achieve the performance of \emph{84.7\% pixel acc.} and \emph{33.2\% mIoU} with a 6-view RGB-input model.

\emph{Multi-modalities VPN.} We demonstrate the results of multi-modalities VPN in Table~\ref{tab:vpn-cross} to show that our VPN can effectively synthesize information from multiple modalities. We set the late-fusion baseline to compare with our multi-modalities VPN, which simply averages the softmax outputs of each single-modality VPN to obtain the final results. We find that the Depth-Semantic VPN achieves the best performance and makes a great improvement. This may be because semantic mask and depth map are two complementary information. However, the Semantic-RGB combination does not bring too much improvement. The reason can be that, for this cross-view semantic segmentation task, semantic input contains most of the useful information in the RGB.

\emph{Importance of View Transformer Module.}
We further evaluate our model in Table~\ref{tab-vpn-abl} to show the importance of the view transformer module. The baseline network is a classic encoder-decoder architecture used in the standard semantic segmentation, in which the encoder and the decoder are the same as our VPN. 
It simply sums up the feature maps from different views and then feeds it to the decoder. Our VPN easily outperforms the baseline and, in some multi-view cases, the baseline model does even worse than single-view one due to the bad fusion strategy.

\emph{Comparing with baseline.}
Table~\ref{tab-vpn} shows that our VPN can easily outperform the 3D geometric method. 3D Geometric method is very easy to fail when there are obstacles. 
In Fig.~\ref{fig:3d_vpn}, we can see that the 3D geometric method is unable to reconstruct the objects which can not be directly observed, even after filling the holes, such as the desk behind the chairs shown in the figure. As for X-Fork, we can see that the original generator performs badly in our cross-view semantic segmentation task. This is because X-Fork doesn't have a necessary module to transform the first-view feature map into the top-down-view space. The ablation study in Table~\ref{tab-vpn-abl} shows a similar issue that there is a significant performance drop when VPN doesn't contain the VTM.

\begin{figure*}
\begin{center}
\includegraphics[width=\linewidth]{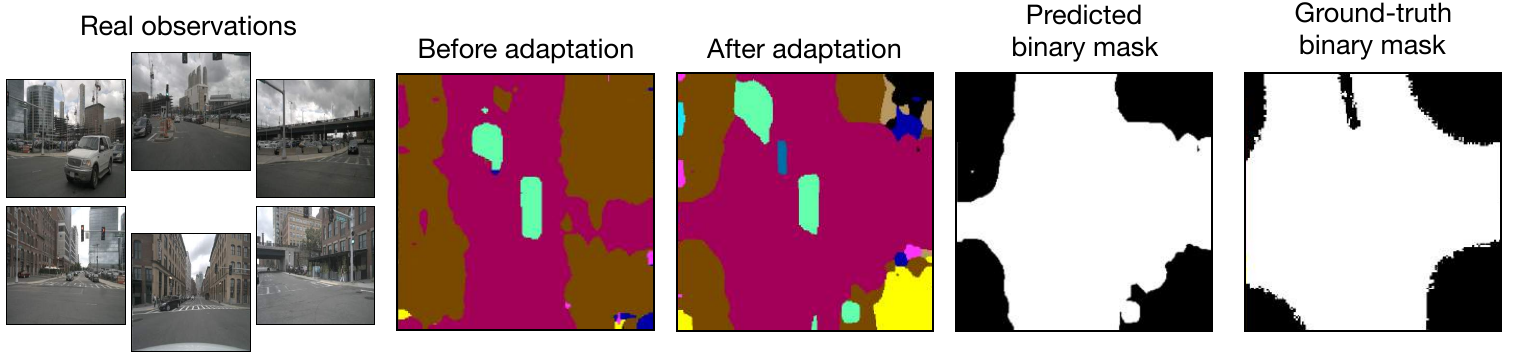}
\end{center}
  \caption{Qualitative results of sim-to-real adaptation. The results of source prediction before and after domain adaptation, drivable area prediciton after adaptation and the groud-truth drivable area map.}
\label{fig:sim_to_real}
\end{figure*}

\subsection{Results of sim-to-real adaptation}\label{sec:sim2real}
After we train and test our VPNs in the simulation environment, we transfer our model to the real-world data. 
We first train a 6-view semantic VPN model on the predicted semantic masks in CARLA simulator and then transfer it to nuScenes dataset by using an unsupervised domain adaptation process as depicted in Section~\ref{sec:syn2real}. We provide the qualitative results in Fig.~\ref{fig:sim_to_real}, from which we can see that our VPN can roughly segment various road shapes like crossroads and also sketch the relative locations of surrounding objects such as cars and buildings. 
As shown in Table~\ref{tab:real-adapt}, we evaluate the quantitative results of real-world performance by using binary drivable-area ground truth. 

\begin{table}[!htb]
\centering
\small
\caption{Results in real world.}
\begin{tabular}{cccc}
    \toprule
Method &  Pix. Acc. & Mean Acc. & mIoU  \\
     \cmidrule(lr){1-4} 
      Before Adaptation & 72.6\% & 61.4\% & 28.0\% \\
     \cmidrule(lr){1-4}
      After Adaptation & $\bm{78.8\%}$ & $\bm{65.2\%}$ & $\bm{31.9\%}$ \\
\bottomrule
\end{tabular}
    \label{tab:real-adapt}
\end{table}

\section{EXPLORATION WITH TOP-DOWN-VIEW MAP}\label{sec:nav_exp}
When exploring an unknown space our humans head to the regions which they have not visited. This intuition reflects that exploration requires the agent to identify free space as well as remember which areas it has not visited yet. To achieve this goal, we make the agent able to identify the free space by training it to predict the top-down-view free-space map. 

\emph{Top-down-view free-space map.}
We train the VPN to predict top-down-view free-space map. Different from the semantic map, free-space map has only two categories, obstacle and free space, which are denoted by 0 1 respectively.

\emph{State map.}
Due to the ideal assumption made above, by memorizing the previous actions it has executed, the agent can easily build the state map which contains the information of the already-visited positions. We label the unvisited pixels as 0 and the already-visited pixels as 1 on the state map.

\emph{Exploration algorithm.}
We detail the navigation policy decision algorithm in Algorithm~\ref{alg:policy}. At each time step $t$, we make the action $a_t$ and update the agent with the next top-down-view free-space map $T_{t+1}$ and state map $S_{t+1}$. In both the top-down-view free-space map and the state map, we assume that the agent is always at the center of the map.
\begin{algorithm}
\caption{Exploration decision policy at time $t$}
\label{alg:policy}
\hspace*{0.02in} {\emph{Input}:}
A top-down-view free-space map $T_t$ and a state map $S_t$ at time step $t$, where $T_t, S_t \in \{0, 1\}^{L\times L}$.  \\
\hspace*{0.01in} {\emph{Output}:}
Policy action $a_t$, where $a_t\in \{$Forward, Back, Left-forward, Right-forward, Done$\}$.
\begin{algorithmic}[1]
\State{$U_t \leftarrow T_t \bigcap \neg S_t$; $a_t \leftarrow$ Done; $d_s \leftarrow + \infty$}
\State{$D_t \leftarrow$ computeDistMap($U_t$)}
\For{$a$ in \{Forward, Back, Left-forward, Right-forward\}}
        \State{$d = $execute($a$)}
        \If{$d_s > d$}
            \State{$d_s \leftarrow d$; $a_t \leftarrow a$}
        \EndIf
\EndFor \\
{\emph{return}} $a_t$
\end{algorithmic}
computeDistMap(): Compute the shortest distance of each map pixel to the unvisited free-space region.\\
execute(): Return the shortest distance of the pixel to which the agent transit if execute the action $a$.
\end{algorithm}

\subsection{Result and comparison}
To demonstrate that VPN can help navigation, we compare it with the following baselines for exploration. \emph{Random walk:} Random walk agent randomly chooses one action from Forward, Back, Right-forward and Left-forward, at each time step. \emph{Top-down-view navigation with ground truth (GT)}: By planning on the ground truth top-down-view free-space map with Algorithm~\ref{alg:policy}, we can obtain the upper-bound performance of our method. The difference is that in our case the top-down-view free-space map is predicted by VPN, rather than the ground truth.  \emph{Imitation learning (IL) without top-down-view:} A reactive CNN network learns to imitate the expert exploration trajectories given the first-view observations.  The trajectories are generated by the baseline above with a top-down-view ground truth map. Network inputs are 4 first-view depth images. 
We also input the state map to indicate the already-visited area. We extract 729 trajectories for the training set and 121 trajectories for the validation set to train the navigation agent. Each trajectory contains 150 states which are all labeled with expert policy.


\begin{table}[!htb]
\centering
\small
\caption{Comparison on exploration.}
\begin{tabular}{cc}
    \toprule
Method &  Coverage Area  \\
     \cmidrule(lr){1-2} 
      Random walk & 260.3 $\pm$ 82.7 \\
     \cmidrule(lr){1-2}
      IL w/o top-down-view & 443.8 $\pm$ 340.6\\
     \cmidrule(lr){1-2}
     Top-down-view navigation & 673.8 $\pm$ 349.8\\
     \cmidrule(lr){1-2}
      Top-down-view navigation with GT &  1070.8$\pm$326.2 \\
\bottomrule
\end{tabular}
    \label{tab:navigation}
\end{table}
\begin{figure}
\begin{center}
\includegraphics[width=\linewidth]{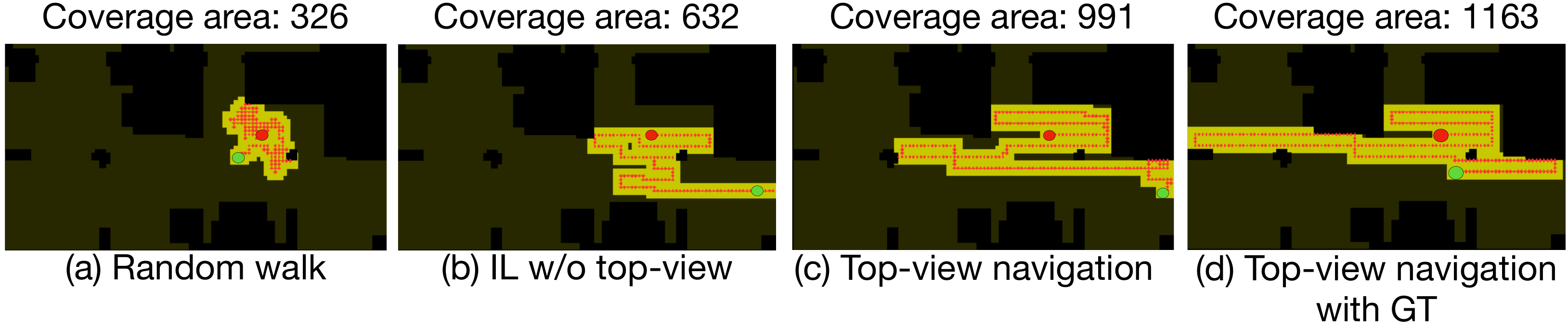}
\end{center}
  \caption{Examples of the surrounding exploration. Start point and end point are marked as red point and green point respectively} 
\label{fig:navigation}
\end{figure}

We run the algorithm directly on our predicted top-down-view map. For testing all the methods, we start the episode by initializing the state maps from zero, indicating that all free space is yet to be visited. \emph{Coverage Area} is defined to measure exploration performance. We randomly choose 100 starting points on a scene map. For each starting point, we let the agent explore the space for 300 steps and compute the coverage area. Then final results are obtained by averaging the coverage area of these 100 episodes. Table~\ref{tab:navigation} plots the exploration result for different methods and Fig.~\ref{fig:navigation} shows some sample trajectories. We can see that equipped with the predicted top-down-view map from our VPN, the agent can efficiently explore the environment. 

\begin{figure}
\begin{center}
\includegraphics[width=\linewidth]{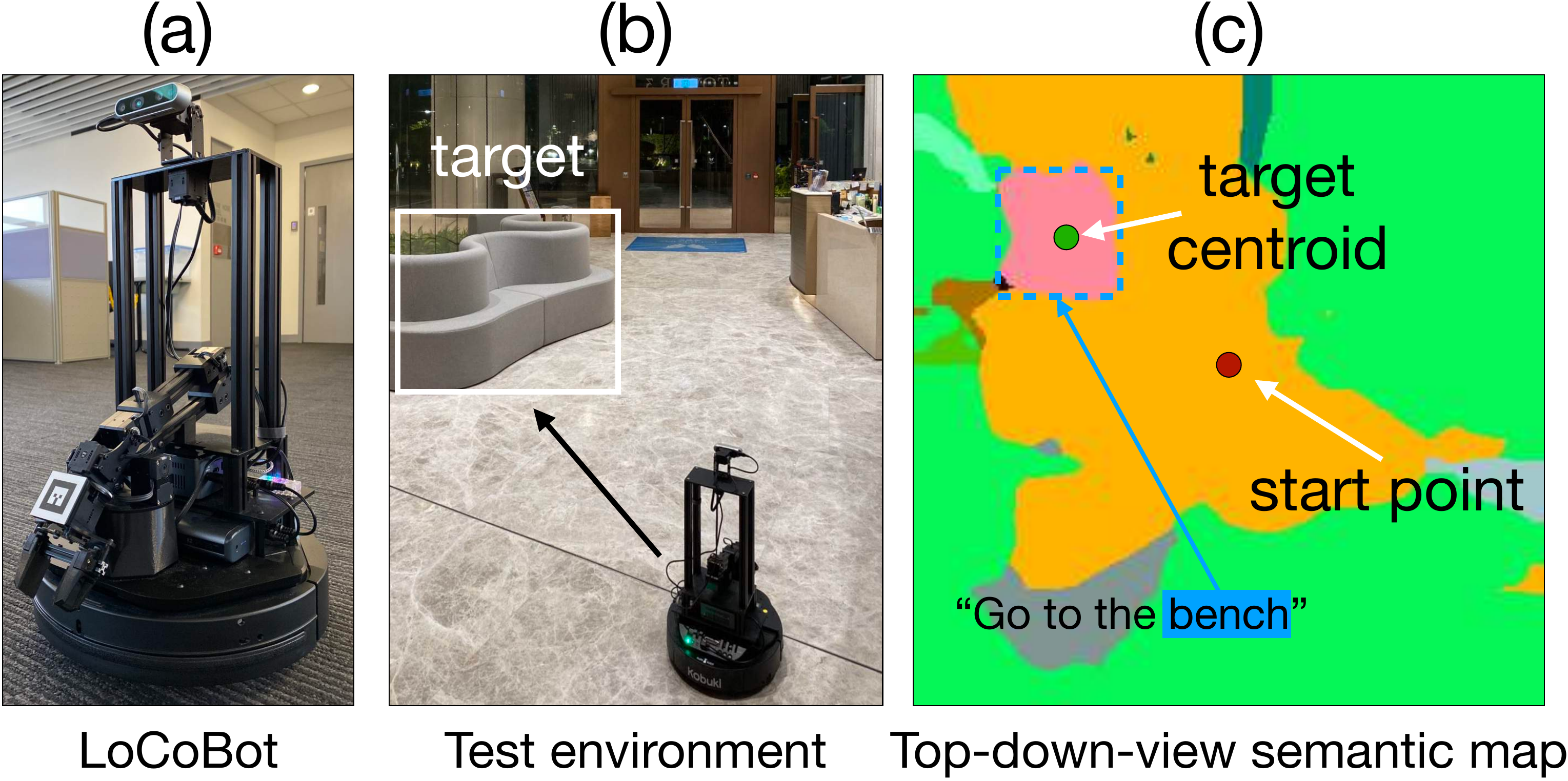}
\end{center}
   \caption{Our experiments are conducted on a LoCoBot mobile robot in the PyRobot platform. (a) We show a picture of the LoCoBot robot. (b) We show our test environment and the target specified by a semantic token (e.g. bench). (c) We show the process that parses the instruction and calculate the target coordinates.}
\label{fig:exp_robot}
\end{figure}

\section{REAL ROBOT EXPERIMENT}\label{sec:real-robot}
To verify the performance of our model in the real-world robotic environment, we conduct a semantic navigation experiment by using a LoCoBot mobile robot~\cite{pyrobot2019} (cf. Fig.~\ref{fig:exp_robot}a). In this task, the robot is required to identify and reach the target specified by a semantic token. For instance, given the instruction ``go to the bench'', the robot has to move to the target which is shown in Fig.~\ref{fig:exp_robot}b. Similar settings are also used in \cite{shen2019situational}. At the initial location, the robot takes 8 RGB images (45 degrees apart) using its head camera. Then it uses the existing semantic segmentation technique to obtain the semantic mask of each RGB image. After that, it predicts the top-down-view semantic map with our VPN. Finally, it parses the instruction and calculates the centroid coordinates of all ``bench'' pixels (cf. Fig.~\ref{fig:exp_robot}c). 
Our real robot experiment shows that though the model is trained in a simulator it exhibits reasonable robustness when we randomly set the initial location and change the layout of surrounding objects. 

\section{CONCLUSION}
In this work, we propose the cross-view semantic segmentation task to sense the environment and a neural architecture design View Parsing Network (VPN) to address that. Based on the experimental results, we demonstrate that VPN can be applied to mobile robots to facilitate the surrounding awareness through a lightweight and efficient top-down-view semantic map. In many situations where the height information of objects is not essential, VPN could be a good alternative compared to the traditional 3D-based methods which are costly on both data memory and computation.

\ifCLASSOPTIONcaptionsoff
  \newpage
\fi



%
\bibliographystyle{IEEEtran}
\bibliography{root}

\begin{thebibliography}{10}
\providecommand{\url}[1]{#1}
\csname url@samestyle\endcsname
\providecommand{\newblock}{\relax}
\providecommand{\bibinfo}[2]{#2}
\providecommand{\BIBentrySTDinterwordspacing}{\spaceskip=0pt\relax}
\providecommand{\BIBentryALTinterwordstretchfactor}{4}
\providecommand{\BIBentryALTinterwordspacing}{\spaceskip=\fontdimen2\font plus
\BIBentryALTinterwordstretchfactor\fontdimen3\font minus
  \fontdimen4\font\relax}
\providecommand{\BIBforeignlanguage}[2]{{%
\expandafter\ifx\csname l@#1\endcsname\relax
\typeout{** WARNING: IEEEtran.bst: No hyphenation pattern has been}%
\typeout{** loaded for the language `#1'. Using the pattern for}%
\typeout{** the default language instead.}%
\else
\language=\csname l@#1\endcsname
\fi
#2}}
\providecommand{\BIBdecl}{\relax}
\BIBdecl

\bibitem{kostavelis2015semantic}
I.~Kostavelis and A.~Gasteratos, ``Semantic mapping for mobile robotics tasks:
  A survey,'' \emph{Robotics and Autonomous Systems}, vol.~66, pp. 86--103,
  2015.

\bibitem{sunderhauf2016place}
N.~S{\"u}nderhauf, F.~Dayoub, S.~McMahon, B.~Talbot, R.~Schulz, P.~Corke,
  G.~Wyeth, B.~Upcroft, and M.~Milford, ``Place categorization and semantic
  mapping on a mobile robot,'' in \emph{2016 IEEE international conference on
  robotics and automation (ICRA)}.\hskip 1em plus 0.5em minus 0.4em\relax IEEE,
  2016, pp. 5729--5736.

\bibitem{cordts2016cityscapes}
M.~Cordts, M.~Omran, S.~Ramos, T.~Rehfeld, M.~Enzweiler, R.~Benenson,
  U.~Franke, S.~Roth, and B.~Schiele, ``The cityscapes dataset for semantic
  urban scene understanding,'' in \emph{Proc. CVPR}, 2016.

\bibitem{dai2017scannet}
A.~Dai, A.~X. Chang, M.~Savva, M.~Halber, T.~A. Funkhouser, and M.~Nie{\ss}ner,
  ``Scannet: Richly-annotated 3d reconstructions of indoor scenes.'' in
  \emph{Proc. CVPR}, vol.~2, 2017, p.~10.

\bibitem{wu2018building}
Y.~Wu, Y.~Wu, G.~Gkioxari, and Y.~Tian, ``Building generalizable agents with a
  realistic and rich 3d environment,'' \emph{arXiv preprint arXiv:1801.02209},
  2018.

\bibitem{Dosovitskiy17}
A.~Dosovitskiy, G.~Ros, F.~Codevilla, A.~Lopez, and V.~Koltun, ``{CARLA}: {An}
  open urban driving simulator,'' in \emph{Proceedings of the 1st Annual
  Conference on Robot Learning}, 2017, pp. 1--16.

\bibitem{zhao2017pspnet}
H.~Zhao, J.~Shi, X.~Qi, X.~Wang, and J.~Jia, ``Pyramid scene parsing network,''
  in \emph{Proc. CVPR}, 2017.

\bibitem{katsumata2019semantic}
Y.~Katsumata, A.~Taniguchi, Y.~Hagiwara, and T.~Taniguchi, ``Semantic mapping
  based on spatial concepts for grounding words related to places in daily
  environments,'' \emph{Frontiers in Robotics and AI}, vol.~6, p.~31, 2019.

\bibitem{zheng2018learning}
K.~Zheng, A.~Pronobis, and R.~P. Rao, ``Learning graph-structured sum-product
  networks for probabilistic semantic maps,'' in \emph{Thirty-Second AAAI
  Conference on Artificial Intelligence}, 2018.

\bibitem{zou2018layoutnet}
C.~Zou, A.~Colburn, Q.~Shan, and D.~Hoiem, ``Layoutnet: Reconstructing the 3d
  room layout from a single rgb image,'' in \emph{Proc. CVPR}, 2018.

\bibitem{hedau2012recovering}
V.~Hedau, D.~Hoiem, and D.~Forsyth, ``Recovering free space of indoor scenes
  from a single image,'' in \emph{Proc. CVPR}, 2012.

\bibitem{schulter2018learning}
S.~Schulter, M.~Zhai, N.~Jacobs, and M.~Chandraker, ``Learning to look around
  objects for top-view representations of outdoor scenes,'' in
  \emph{Proceedings of the European Conference on Computer Vision (ECCV)},
  2018, pp. 787--802.

\bibitem{wang2019parametric}
Z.~Wang, B.~Liu, S.~Schulter, and M.~Chandraker, ``A parametric top-view
  representation of complex road scenes,'' in \emph{Proceedings of the IEEE
  Conference on Computer Vision and Pattern Recognition}, 2019, pp.
  10\,325--10\,333.

\bibitem{zhai2017predicting}
M.~Zhai, Z.~Bessinger, S.~Workman, and N.~Jacobs, ``Predicting ground-level
  scene layout from aerial imagery,'' in \emph{Proc. CVPR}, vol.~3, 2017.

\bibitem{regmi2018cross}
K.~Regmi and A.~Borji, ``Cross-view image synthesis using conditional gans,''
  in \emph{Proc. CVPR}, 2018, pp. 3501--3510.

\bibitem{shen2019situational}
W.~B. Shen, D.~Xu, Y.~Zhu, L.~J. Guibas, L.~Fei-Fei, and S.~Savarese,
  ``Situational fusion of visual representation for visual navigation,'' in
  \emph{Proceedings of the IEEE International Conference on Computer Vision},
  2019, pp. 2881--2890.

\bibitem{tsai2018learning}
Y.-H. Tsai, W.-C. Hung, S.~Schulter, K.~Sohn, M.-H. Yang, and M.~Chandraker,
  ``Learning to adapt structured output space for semantic segmentation,'' in
  \emph{Proceedings of the IEEE Conference on Computer Vision and Pattern
  Recognition}, 2018, pp. 7472--7481.

\bibitem{sun2019high}
K.~Sun, Y.~Zhao, B.~Jiang, T.~Cheng, B.~Xiao, D.~Liu, Y.~Mu, X.~Wang, W.~Liu,
  and J.~Wang, ``High-resolution representations for labeling pixels and
  regions,'' \emph{arXiv preprint arXiv:1904.04514}, 2019.

\bibitem{nuscenes2019}
H.~Caesar, V.~Bankiti, A.~H. Lang, S.~Vora, V.~E. Liong, Q.~Xu, A.~Krishnan,
  Y.~Pan, G.~Baldan, and O.~Beijbom, ``nuscenes: A multimodal dataset for
  autonomous driving,'' \emph{arXiv preprint arXiv:1903.11027}, 2019.

\bibitem{pyrobot2019}
A.~Murali, T.~Chen, K.~V. Alwala, D.~Gandhi, L.~Pinto, S.~Gupta, and A.~Gupta,
  ``Pyrobot: An open-source robotics framework for research and benchmarking,''
  \emph{arXiv preprint arXiv:1906.08236}, 2019.

\end{thebibliography}




%








\end{document}